\documentclass[12pt]{l4dc2020} 

\makeatletter
\def\set@curr@file#1{\def\@curr@file{#1}} 
\makeatother
\usepackage[load-configurations=version-1]{siunitx} 

\title[Robust Learning-Based Control via Bootstrapped Multiplicative Noise]{Robust Learning-Based Control via Bootstrapped Multiplicative Noise}
\usepackage{times}
\usepackage{algorithm}
\usepackage{algorithmic}

\newcommand{\expname}{1628209502p0156405_Ns_100000_T_200_system_scalar_seed_1}

\author{%
 \Name{Benjamin Gravell} \Email{benjamin.gravell@utdallas.edu}\\
 \addr University of Texas at Dallas
 \AND
 \Name{Tyler Summers} \Email{tyler.summers@utdallas.edu}\\
 \addr University of Texas at Dallas%
}

\begin{document}

\maketitle

\begin{abstract}%
Despite decades of research and recent progress in adaptive control and reinforcement learning, there remains a fundamental lack of understanding in designing controllers that provide robustness to inherent \emph{non-asymptotic} uncertainties arising from models estimated with finite, noisy data. We propose a robust adaptive control algorithm that explicitly incorporates such non-asymptotic uncertainties into the control design. The algorithm has three components: (1) a least-squares nominal model estimator; (2) a bootstrap resampling method that quantifies non-asymptotic variance of the nominal model estimate; and (3) a non-conventional robust control design method using an optimal linear quadratic regulator (LQR) with multiplicative noise. A key advantage of the proposed approach is that the system identification and robust control design procedures both use stochastic uncertainty representations, so that the actual inherent statistical estimation uncertainty directly aligns with the uncertainty the robust controller is being designed against. 
We show through numerical experiments that the proposed robust adaptive controller can significantly outperform the certainty equivalent controller on both expected regret and measures of regret risk.
\end{abstract}


\section{Introduction}

Recent high profile successes and the resulting hype in machine learning and reinforcement learning are generating renewed interest in adaptive control and system identification, which have their own decades-long histories \cite{aastrom2013adaptive,ljung1998system}.
Classical work on adaptive control and system identification largely focused on asymptotics, including stability, consistency, asymptotic variance, etc. Emerging research at the intersection of learning and control shifts focus to non-asymptotic statistical analyses, including regret and sample efficiency in various adaptive control and learning algorithms \cite{abbasi2011regret,dean2018regret,Dean2019}.

In both classical and emerging work in learning and control, robustness has been a key issue. The classical focus on asymptotics led to a strong emphasis on certainty-equivalent adaptive control, where inevitable uncertainty in model estimates is ignored for control design and unsurprisingly can lead to serious lack of robustness. It remains poorly understood how to best interface both asymptotic and non-asymptotic uncertainty descriptions of model estimates with robust control design methods. One of the main difficulties is a mismatch in uncertainty descriptions: system identification almost universally uses stochastic data models\footnote{A notable exception is set membership identification \cite{milanese1991optimal}, which maintains a \emph{set} of models that could have produced the data within some error bound and interfaces somewhat more naturally with certain traditional robust control design methods.}, whereas robust control traditionally uses set-based descriptions and worst-case design. This can lead to unnecessary conservatism when the assumed model uncertainty sets are poorly aligned with inherent statistical model uncertainty. 

The contributions of the present work are as follows:
\begin{enumerate}
    \item We propose a robust adaptive control algorithm where the model uncertainty description and robust control design method both use stochastic uncertainty representations.
    \item We show via numerical experiments that the proposed robust adaptive controller can significantly outperform the certainty equivalent controller on both expected regret and measures of regret risk.
\end{enumerate}
 The algorithm has three components: (1) a least-squares nominal model estimator; (2) a bootstrap resampling method that quantifies non-asymptotic variance of the nominal model estimate; and (3) a non-conventional robust control design method using an optimal linear quadratic regulator (LQR) with multiplicative noise. This approach provides a natural interface between two widely used and highly effective methods from statistics and optimal control theory (namely, bootstrap sample variance and LQR). It is known that certainty equivalent adaptive control can achieve asymptotic optimality and statistical efficiency with an \emph{order} optimal rate \cite{chen2012identification,kumar2015stochastic,mania2019certainty}. However, neither of these imply anything about non-asymptotic optimality or robustness.

\section{Problem Formulation}
We consider adaptive control of the discrete-time linear dynamical system
\begin{align}
    x_{t+1} = A x_t + B u_t + w_t, 
\end{align}
where $x_t \in \mathbf{R}^n$ is the system state, $u_t \in \mathbf{R}^m$ is the control input, and $w_t$ is i.i.d. process noise with zero mean and covariance matrix $W$.  The system matrices $(A,B)$ are assumed unknown, so an adaptive controller is to be designed based only on state-input trajectory data $x_{0:t} := [x_0, ..., x_t]$, $u_{0:t-1} = [u_0,...,u_{t-1}]$. We consider the linear quadratic adaptive optimal control problem
\begin{equation} \label{aoc}
     \min_{\pi}\mathbf{E} \left[ \sum_{t=0}^\intercal (x_t^\intercal Q x_t + u_t^\intercal R u_t - J^*)\right] 
\end{equation}
where $Q \succeq 0$ and $R \succeq 0$ are cost matrices, and the optimization is over (measureable) history dependent feedback policies $\pi = \{ \pi_t \}_{t=0}^{T-1}$ with $u_t = \pi_t(x_{0:t}, u_{0:t-1})$. The constant $J^*$ in the stage costs represents the optimal infinite-horizon average steady state cost when the system matrices $(A,B)$ are known, which results from a static linear state feedback, $u_t = K x_t$, whose gain matrix $K$ can be computed via several known methods, including value iteration, policy iteration, and semidefinite programming. This constant gives the stage cost an interpretation as \emph{regret}.

The finite horizon objective in \eqref{aoc} emphasizes the \emph{non-asymptotic} performance of the adaptive controller. This stands in contrast to a majority of classical work on adaptive control, which tends to focus on asymptotic performance and stability. 
Note that regret is a random variable that ideally should be small, and there are various ways to measure its size, including expected regret, regret variance, and measures of regret risk, such as value at risk or conditional value at risk. 

It has been long known that this problem can be solved \emph{in principle} by redefining the state as the (infinite-dimensional) joint conditional distribution over the original state and unknown model parameters and applying dynamic programming \cite{bellman1961adaptive}. However, this approach is intractable even for the most trivial instances. Since computing the optimal policy exactly appears to be intractable, we instead aim to design a computationally implementable controller with good performance and robustness properties, i.e., one that achieves both small expected regret and small regret risk. In particular, our algorithm accounts for uncertainty in various \emph{directions} by modeling them as multiplicative noises, in contrast to the isotropic robustness afforded by the system-level synthesis in \cite{dean2018regret,Dean2019}.
We compare with a certainty equivalent adaptive controller, where uncertainty is ignored and a controller is designed as if the point model estimates were exact.

\section{Preliminaries: Multiplicative Noise LQR}

Here we will represent parameter uncertainty in model estimates stochastically using covariance matrices estimated from bootstrap resampling of finite data records. This representation interfaces quite naturally with a variant of the linear quadratic regulator that incorporates multiplicative noise, which has a long history in control theory but is far less widely known than its additive noise counterpart \cite{Wonham1967,gravell2019learning}. Consider the linear quadratic regulator problem with dynamics perturbed by multiplicative noise
\begin{alignat}{2}  \label{eq:mLQR}
    &\underset{{\pi \in \Pi}}{\text{minimize}} \quad && \mathbf{E}_{x_0,\{\bar A_t\}, \{\bar B_t \}} \sum_{t=0}^\infty (x_t^\intercal Q x_t + u_t^\intercal R u_t), \\
    &\text{subject to}                         \quad && x_{t+1} = (A  + \bar A_t) x_t +  (B + \bar B_t) u_t, \nonumber
\end{alignat}
where $\bar A_t$ and $\bar B_t$ are i.i.d. zero-mean random matrices with a joint covariance structure over their entries governed by the covariance matrices $\Sigma_A := \mathbf{E} [\mathbf{vec}(\bar A)\mathbf{vec}(\bar A)^\intercal ] \in \mathbf{R}^{n^2 \times n^2}$ and  $\Sigma_B := \mathbf{E} [ \mathbf{vec}(\bar B)\mathbf{vec}(\bar B)^\intercal ] \in \mathbf{R}^{nm \times nm}$, which quantify uncertainty in the nominal system matrices $(A,B)$ and can be estimated from data using bootstrap methods.

Just as in additive noise LQR problems, dynamic programming can be used to show that the optimal policy is linear state feedback $u_t = K x_t$. Given the problem data $(A,B,Q,R,\Sigma_A, \Sigma_B)$, the optimal quadratic cost matrix is given by the solution of the \emph{generalized} Riccati equation
\begin{align} \label{eq:genriccati}
    P   & = Q + A^\intercal P A + \sum_{i=1}^{n^2} \alpha_i A_i^\intercal P A_i - A^\intercal P B (R + B^\intercal P B + \sum_{j=1}^{nm} \beta_j B_j^\intercal P B_j)^{-1} B^\intercal P A,
\end{align}
and the associated optimal gain matrix is $K = -  \big(R + B^\intercal P B + \sum_{j=1}^{nm} \beta_j B_j^\intercal P B_j \big)^{-1} B^\intercal P A$, where $\{\alpha_i, A_i\}_{i=1}^{n^2}$ and $\{\beta_j, B_j\}_{j=1}^{nm}$ are the eigenvalues and reshaped eigenvectors of $\Sigma_A$ and $\Sigma_B$, respectively.
The solutions are denoted $(P,K) = \text{GDARE}(A, B, Q, R, \Sigma_A, \Sigma_B)$.
The optimal cost and policy can be computed via value iteration, policy iteration, or semidefinite programming (\cite{ElGhaoui1995,gravell2019learning}). Note that like traditional robust control but unlike additive noise LQR, the optimal cost matrix and control gain depend explicitly on the model uncertainty.

\section{Robust Adaptive Control via Bootstrapping and Multiplicative Noise}
Our robust adaptive control algorithm is summarized in Figure \ref{fig:block_diagram} and Algorithm \ref{algorithm:algo1}. The algorithm has three main components: (1) a least-squares nominal model estimator; (2) a bootstrap resampling method that quantifies non-asymptotic variance of the nominal model estimate; and (3) a non-conventional robust control design method using an optimal LQR with multiplicative noise.

\begin{figure}[!htbp]
\floatconts
  {fig:block_diagram}
  {\vspace{-0.5cm}\caption{Block diagram of our robust adaptive control algorithm.}}
  {%
      \includegraphics[width=0.7\linewidth]{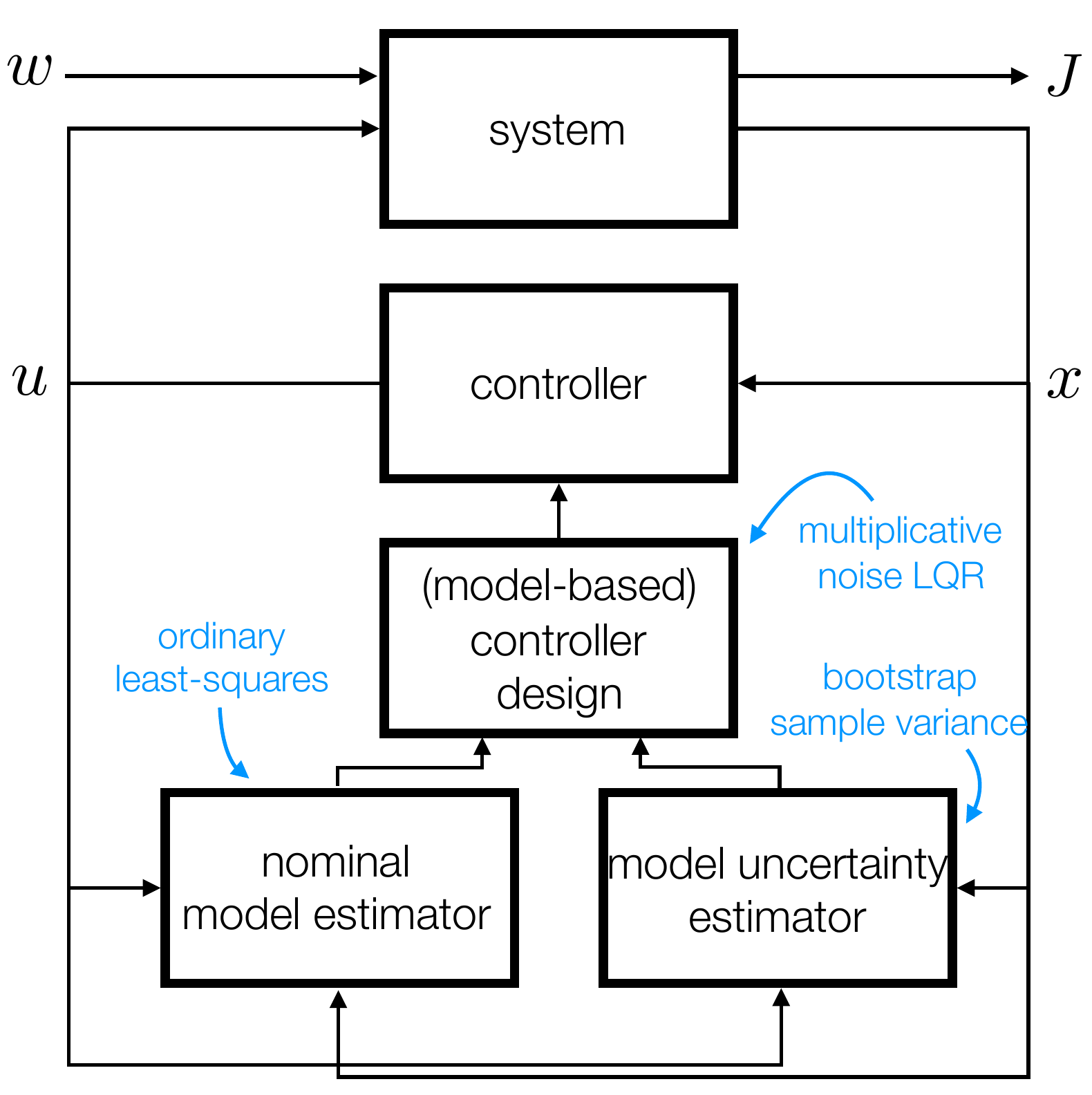}
  }
\end{figure}

\begin{algorithm}
\caption{Robust Adaptive Control}
\begin{algorithmic}[1]
\label{algorithm:algo1}
    \REQUIRE exploration time $t_{\text{explore}}$, input excitation covariance $U$, number of bootstrap resamples $N_b$, model uncertainty scaling parameter $\gamma$, cost matrices $Q \succeq 0, R \succeq 0$
    \STATE $x_0 \sim \mathcal{N}(0, X_0)$
    \FOR{$t=0,\ldots,T-1$}
    	\IF{$t \leq t_{\text{explore}}$}
	\STATE $u_t  \sim \mathcal{N}(0, U)$
	\STATE $x_{t+1} = A x_t + B u_t + w_t$
	\ELSE 
	\STATE $(\hat A_t, \hat B_t) = \tt{OrdinaryLeastSquares}$$(x_{0:t}, u_{0:t-1})$
	\STATE $(\hat \Sigma_{A_t}, \hat \Sigma_{B_t}) = \tt{BootstrapModelVariance}$$(x_{0:t}, u_{0:t-1})$
	\STATE $\hat K_t = \tt{MultiplicativeNoiseLQR}$$(\hat A_t, \hat B_t, Q, R, \gamma \hat \Sigma_{A_t}, \gamma \hat \Sigma_{B_t})$
	\STATE $e_t \sim \mathcal{N}(0,  \|(\hat \Sigma_{A_t}, \hat \Sigma_{B_t}) \| U)$
	\STATE $u_t = \hat K_t x_t + e_t$
	\STATE $x_{t+1} = A x_t + B u_t + w_t$
	\ENDIF
    \ENDFOR
\end{algorithmic}
\end{algorithm}

\subsection{Least Squares Estimation for the Nominal Model}
The first component of the algorithm is a standard least-squares estimator for the unknown system matrices from state-input trajectory data. In particular, at time $t$ from data $(x_{0:t}, u_{0:t-1})$, we form the estimate
\begin{equation}
    (\hat A_t, \hat B_t) = {\text{argmin}}_{(A, B)} \left\{ \sum_{\tau=0}^{t-1}  \|x_{t+1} - (A x_t + B u_t) \|_2^2 \right\}.
\end{equation}
More explicitly, defining the data matrices $X_t^\intercal = [x_1 \ \ x_2 \ \ \cdots \ \ x_t]$, $Z_t^\intercal = \left[\begin{array}{cccc}x_0 & x_1 & \cdots & x_{t-1} \\u_0 & u_1 & \cdots & u_{t-1}\end{array}\right]$
then the least squares estimate can be written as
\begin{equation} \label{eq:lse}
    [\hat A_t, \  \hat B_t] = X_t^\intercal Z_t (Z_t^\intercal Z_t)^{-1}.
\end{equation}
A non-degenerate model estimate is obtained only when $Z_t^\intercal Z_t$ is invertible, so learning is divided into a pure exploration phase until a user-specified time $t_{\text{explore}} > n+m$ and subsequently an exploration-exploitation phase, where the estimated model is used to design a control policy. The exploration component of the input signal is iid Gaussian noise with user-specified covariance matrix $U$; it is well known that for any $U \succ 0$ the least-squares estimator is consistent under our modeling assumptions. The exploration noise is designed to fade out with the bootstrap-estimated model uncertainty, which yields asymptotic optimality. Under mild assumptions, the least-squares estimator can be implemented recursively to significantly simplify the repeated computation in \eqref{eq:lse} as data arrives (e.g. \cite{simon2006optimal}).

\subsection{Bootstrap Resampling to Quantify Non-Asymptotic Model Uncertainty}
There are inevitably errors in the least-squares estimate obtained from any finite data record, due to the process noise affecting the system dynamics. Due to dependence in the time series data, unfortunately it is not straightforward to analytically characterize non-asymptotic uncertainty in the least-squares estimate using standard statistical techniques. Therefore, to quantify non-asymptotic uncertainty in the model estimate, we propose a time series bootstrap resampling procedure. There are three broad bootstrap methods for time series \cite{hardle2003bootstrap}: (1) Parametric bootstrap; (2) Semi-parametric bootstrap with resampled residuals; (3) Non-parametric bootstrap with block resampling. In parametric and semi-parametric methods, bootstrap data are simulated from the nominal model with the process noise sampled iid with replacement either from an assumed distribution or from residuals calculated with the nominal model. Dependence in the data is preserved by construction. In non-parametric methods, overlapping time blocks of consecutive data are sampled from the original data to preserve dependence. For definiteness, a semi-parametric bootstrap with resampled residuals for the least-squares estimator discussed above is summarized in Algorithm \ref{algorithm:algo2}.
\begin{algorithm}
\caption{Semi-Parametric Bootstrap}
\begin{algorithmic}[1]
\label{algorithm:algo2}
    \REQUIRE trajectory data $(x_{0:t}, u_{0:t-1})$, nominal model estimate $(\hat A_t, \hat B_t)$, residuals $\hat w_\tau= x_{\tau + 1} - (\hat A_t x_\tau + \hat B_t u_\tau)$, $\tau = 0,...,t-1$, number of bootstrap resamples $N_b$
    \STATE $\bar x_0 = x_0$
    \STATE $\bar u_{0:t-1} = u_{0:t-1}$ \;
    \FOR{$k=1,\ldots,N_b$}
        \STATE Generate data $\bar x_{\tau+1} = \hat A_t \bar x_\tau + \hat B_t \bar u_\tau  + \tilde w_\tau$, $\tau=0,...,t-1$, where $\{ \tilde w_\tau \}_{\tau=0}^{t-1}$ is an iid resample with replacement from residuals $\{ \hat w_\tau \}_{\tau=0}^{t-1}$  \;
        \STATE $(\hat A_t^k, \hat B_t^k) = {\text{argmin}}_{(A, B)} \left\{ \sum_{\tau=0}^{t-1}  \| \bar x_{\tau+1} - (A \bar x_\tau + B \bar u_\tau) \|_2^2 \right\}$
    \ENDFOR
    \ENSURE Bootstrap sample covariance $\hat \Sigma_{A_t} = \frac{1}{N_b-1} \sum_{k=1}^{N_b} \text{vec}(\hat A_t^k - \hat A_t) \text{vec}(\hat A_t^k - \hat A_t)^\intercal$ \\
    \qquad \ Bootstrap sample covariance $\hat \Sigma_{B_t} = \frac{1}{N_b-1} \sum_{k=1}^{N_b} \text{vec}(\hat B_t^k - \hat B_t) \text{vec}(\hat B_t^k - \hat B_t)^\intercal$ \;
\end{algorithmic}
\end{algorithm}

\subsection{Multiplicative Noise LQR}
The least squares estimator and boostrap provide both a nominal estimate of the system model and an estimate of the covariance of the nominal model error. These quantities provide precisely the input data needed to compute an optimal policy for the LQR problem with multiplicative noise from the generalized Riccati equation \eqref{eq:genriccati}. This policy is known to provide robustness to uncertainties in the parameters of the nominal model \cite{bernstein1986robust}. Furthermore, the uncertainty in the nominal model estimate used in this control design method is richly structured and derived directly from the finite available data.

We introduce a parameter $\gamma$ which provides a fixed scaling of the model uncertainty. Note that $\gamma = 0$ corresponds to certainty equivalent adaptive control, and as $\gamma$ increases, more weight is placed on uncertainty in the nominal model. Existence of a solution to the generalized Riccati equation \eqref{eq:genriccati} depends not just on stabilizability of the nominal system $(A,B)$, but also on the \emph{mean-square stabilizability} of the multiplicative noise system. When the multiplicative noise variances are too large, it may be impossible to stabilize the system in the mean-square sense.
In this case, we scale down the model variances at each time step if necessary to compute a mean-square stabilizing control gain via bisection; see Algorithm \ref{algorithm:algo3}.
 
In particular, we verify the system with specified $\gamma$ is mean-square stabilizable by checking whether the generalized Riccati equation in \eqref{eq:genriccati} admits a positive semidefinite solution; if not, we find the upper limit $\gamma_\text{max} = c_\gamma \gamma$ via bisection (e.g. \cite{burden1978numerical}) on a scaling $c_\gamma \in [0,1]$.

\begin{algorithm}
\caption{Multiplicative Noise LQR}
\begin{algorithmic}[1]
\label{algorithm:algo3}
    \REQUIRE Nominal model matrices $A$, $B$, cost matrices $Q$, $R$, multiplicative noise scaling $\gamma$ and covariances $\hat{\Sigma}_A$, $\hat{\Sigma}_B$, bisection tolerance $\epsilon > 0$
    \STATE Find largest $c_{\gamma} \in [0,1]$ via bisection such that there exists a feasible solution to \eqref{eq:genriccati} \\
    $(P, K) = \text{GDARE}(A, B, Q, R, c_{\gamma} \gamma \Sigma_A, c_{\gamma} \gamma \Sigma_B)$ \;
    \ENSURE Cost matrix $P$, gain matrix $K$ \;
\end{algorithmic}
\end{algorithm}

\section{Numerical Experiments}

For brevity, we abbreviate ``certainty-equivalent'' as ``CE'' and ``robustness via multiplicative noise'' as ``RMN''.
To evaluate the performance of the proposed RMN algorithm relative to CE control, we performed Monte Carlo sampling to estimate the distribution of several key quantities: instantaneous regret, model error, and multiplicative noise variances.

The instantaneous regret is heavy-tailed in the sense that the effect of outliers with non-negligible probability is significant: some exceptionally poor sequences of model estimates induce extremely high costs relative to the median. For this reason, to facilitate the most direct comparison of CE and the proposed RMN approach, we train the models using both control schemes on identical offline training data. This way the effect of outlier model estimates is applied uniformly to both algorithms, since at each time the algorithms are faced with exactly the same model estimates. In an online adaptive control setting, where the training data are the actual state trajectories experienced under adaptive control, a direct comparison of the approaches is more difficult. While we observe qualitatively similar benefits of the proposed approach, our future work will study this setting with a greater number of Monte Carlo samples to reduce the effect of outliers.

The training data are generated by initializing the state at the origin, applying random controls distributed according to a standard Gaussian distribution (zero-mean, identity-covariance), and simulating the evolution of the state with the additive process noise specified by the problem data $W$. The resulting training data are a set of state trajectories $x_{0:t}^{(k)}$ and input trajectories $u_{0:t-1}^{(k)}$.
Model and uncertainty estimates are generated at time $t$ according to Algorithm \ref{algorithm:algo1} using training data only up to time $t$. 
The optimal cost is empirically calculated by averaging over all Monte Carlo samples the cost incurred by trajectories under optimal control $u_t=K^* x_t$, $K^* = \text{DARE}(A, B, Q, R)$ for each additive noise realization, i.e.
\begin{align*}
    c^*_t &= \frac{1}{N_s} \sum_{k=0}^{N_s} {x_t^{*,(k)}}^\intercal Q {x_t^{*,(k)}} + {u_t^{*,(k)}}^\intercal R {u_t^{*,(k)}} \\
    \text{where } {x_{t+1}^{*,(k)}} &= A {x_t^{*,(k)}} + B {u_t^{*,(k)}} + {w_t^{(k)}}, \ {u_t^{*,(k)}} = K^* {x_t^{*,(k)}}
\end{align*}
The empirical cost under adaptive control $c_t^{(k)}$ is calculated similarly without averaging over Monte Carlo samples.
Instantaneous regret is calculated by subtracting the empirical cost under optimal control from the empirical cost under adaptive control using each scheme i.e. $r_t^{(k)}=c_t^{(k)}-c^*_t$.

We evaluated the CE and RMN algorithms on a scalar system with true system and cost parameters
${A = 1}$, 
${B = 1}$, 
${Q = 1}$, 
${R = 0}$, 
${W = 1}$.
The level of additive process noise is significant enough that an appreciable number of model estimates remain poor for many timesteps; this is necessary to observe a difference between CE and RMN control. We simulated the system over a time horizon of $T = 200$ steps.
We drew $N_{s} = 100,000$ independent Monte Carlo samples and $N_{b} = 100$ bootstrap samples at each time step for uncertainty estimation.
We used unity scaling of the multiplicative noise ($\gamma = 1$) and a tolerance of $\epsilon = 0.01$ for bisection to find the largest scaling $c_\gamma$ of multiplicative noise variance in the multiplicative noise LQR algorithm.
We used an exploration time of $t_\text{explore} = 5$ which ensures the least-squares estimate is non-degenerate. The figures have x-axis limits truncated to $[t_\text{explore}, T]$.

In Figure \ref{fig:plot_instant_regret} we plot statistics of instantaneous regret using CE control and using RMN control.
We are chiefly interested in performance in terms of expected regret and upper quantiles, which correspond to regret risk.
Figure \ref{fig:plot_instant_regret} demonstrates that the multiplicative noise control achieves much lower instantaneous regret in terms of both the mean and upper quantiles. 
In particular, we see that the performance of the multiplicative noise control is clearly many orders of magnitude better between the start and $t=80$, $t=40$, $t=20$ for the 99.9th, 99th and 95th quantiles, respectively. After several time steps the model estimates improve and uncertainty estimates become sufficiently small that the difference between CE and RMN control becomes insignificant. The heaviness of regret tails is shown by the massive difference between the mean and median.

In Figure \ref{fig:plot_ABerr} we plot statistics of the nominal model estimate errors, which are applicable to both control schemes. This shows that the least-squares estimator provides models of increasing accuracy as time goes on, as expected. In Figure \ref{fig:plot_alphabeta} we plot statistics of the multiplicative noise variances using RMN control. This shows that the multiplicative noise variances accurately reflect the true model error, i.e., the boostrap model uncertainty estimator gives reasonable estimates.

The benefits of RMN control over CE control more generally obviously cannot be inferred from this single instance. Indeed other preliminary numerical results on higher dimensional examples indicate that on some problem data ($A$, $B$, $Q$, $R$, $W$) the benefits of RMN control and how best to select the algorithm parameters are unclear, especially in initial stages when there is very high uncertainty around the nominal model estimates. However, there are at least some systems, like the one shown here, that are controlled with significantly lower risk using RMN control. We expect that by explicitly incorporating model uncertainty into the adaptive control design, it should be possible to realize the observed robustness benefits more broadly, which motivates further theoretical study.
\noindent
Code which realizes the algorithms of this paper and generates the reported results is available from \\
\url{https://github.com/TSummersLab/robust-adaptive-control-multinoise}.

\begin{figure}[htbp]
\floatconts
  {fig:plot_instant_regret}
  {\vspace{-0.5cm}\caption{Instantaneous regret vs time for the example system using using certainty-equivalent (a) and multiplicative noise (b) control.}}
  {%
    \subfigure[CE]{\label{fig:plot_instant_regret_ce}%
      \includegraphics[width=0.47\linewidth]{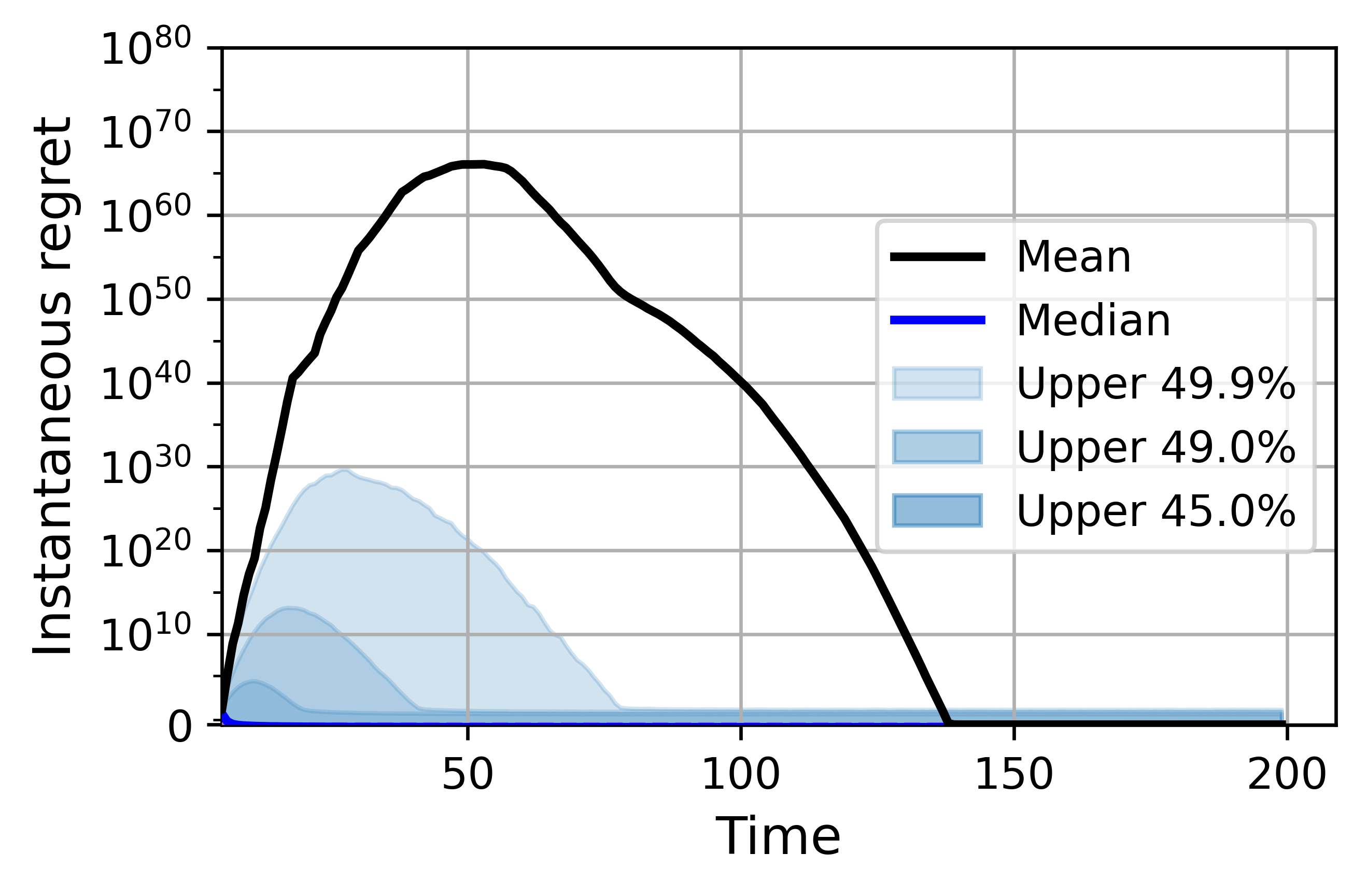}}%
    \qquad
    \subfigure[RMN]{\label{fig:plot_instant_regret_rmn}%
      \includegraphics[width=0.47\linewidth]{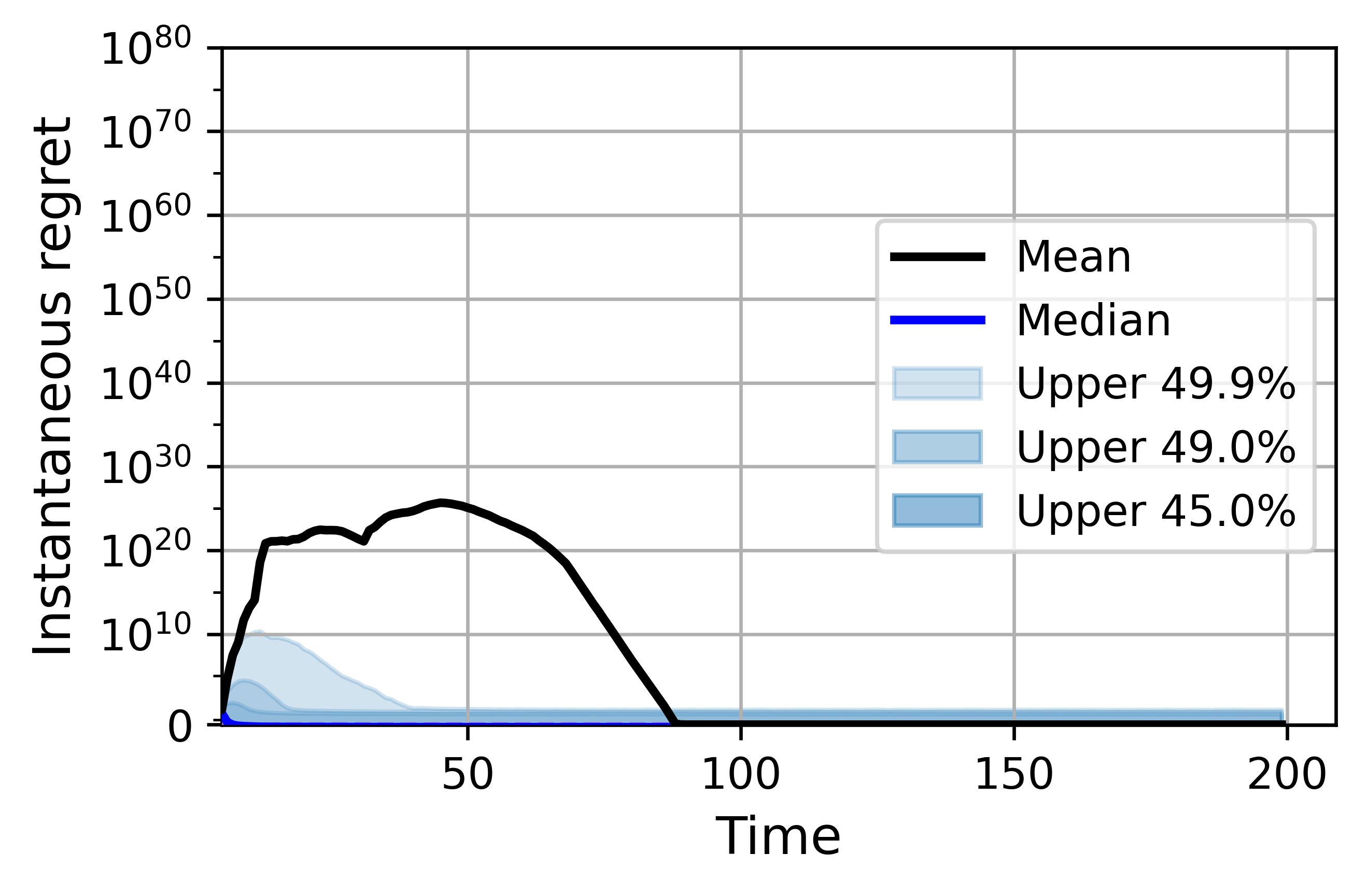}}
  }
\vspace{-0.5cm}
\end{figure}



\section{Conclusions}

We proposed a robust adaptive control algorithm that uses the bootstrap to estimate model estimate covariances and a non-conventional multiplicative noise LQR robust control method. Ongoing and future work will go towards providing finite-time theoretical performance guarantees using tools from high-dimensional statistics,
finding algorithm parameters that ensure uniform improvements over certainty-equivalent control for any system,
and implementing model uncertainty estimates using recursive least-squares to alleviate computational burden.

\acks{This material is based on work supported by the United States Air Force Office of Scientific Research under award number FA2386-19-1-4073.}

\begin{figure}[htbp]
\floatconts
  {fig:plot_ABerr}
  {\vspace{-0.5cm}\caption{Absolute error in estimated (a) A and (b) B matrices vs time for the example system.}}
  {%
    \subfigure[$A$]{\label{fig:plot_Aerr}%
      \includegraphics[width=0.42\linewidth]{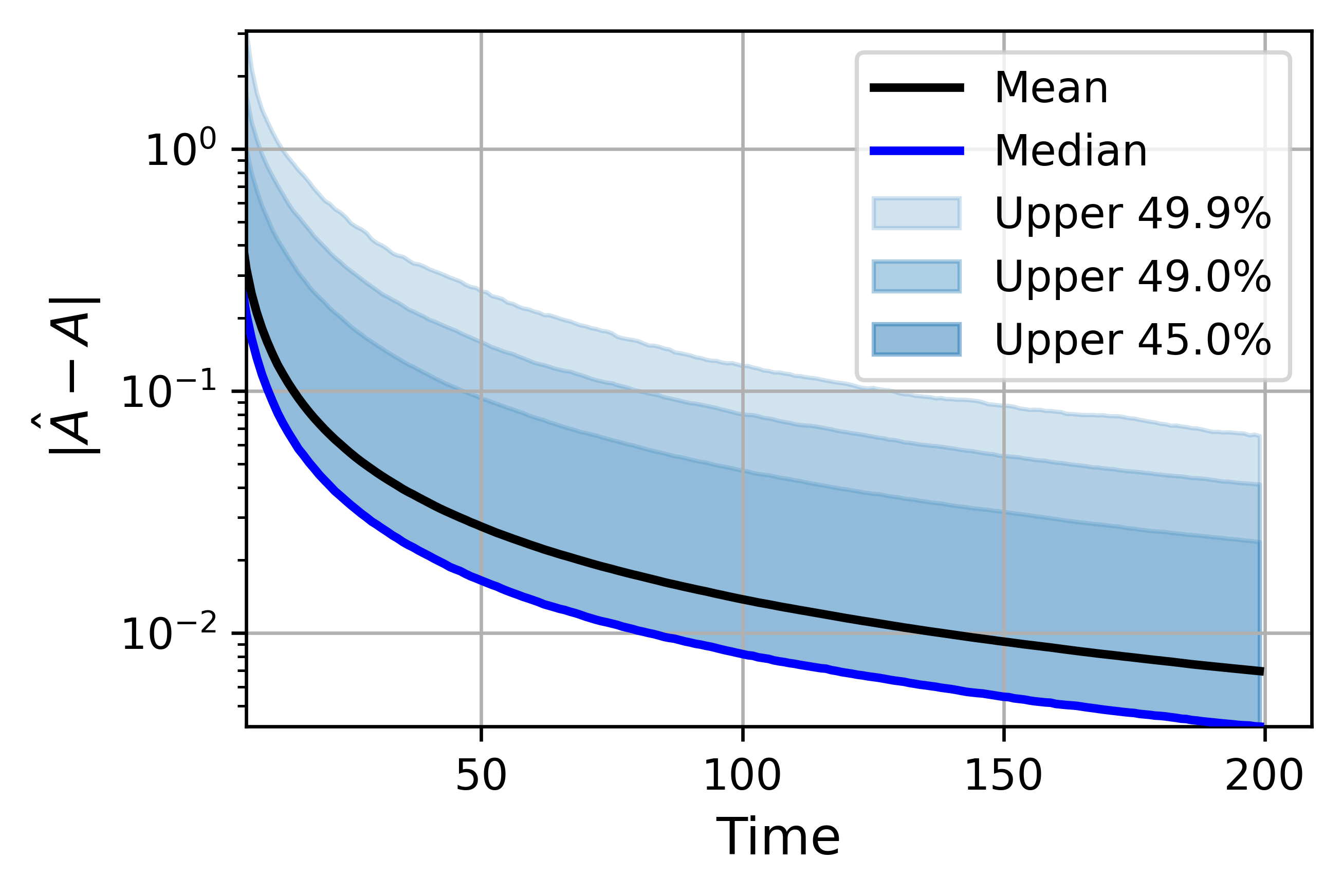}}%
    \qquad
    \subfigure[$B$]{\label{fig:plot_Berr}%
      \includegraphics[width=0.42\linewidth]{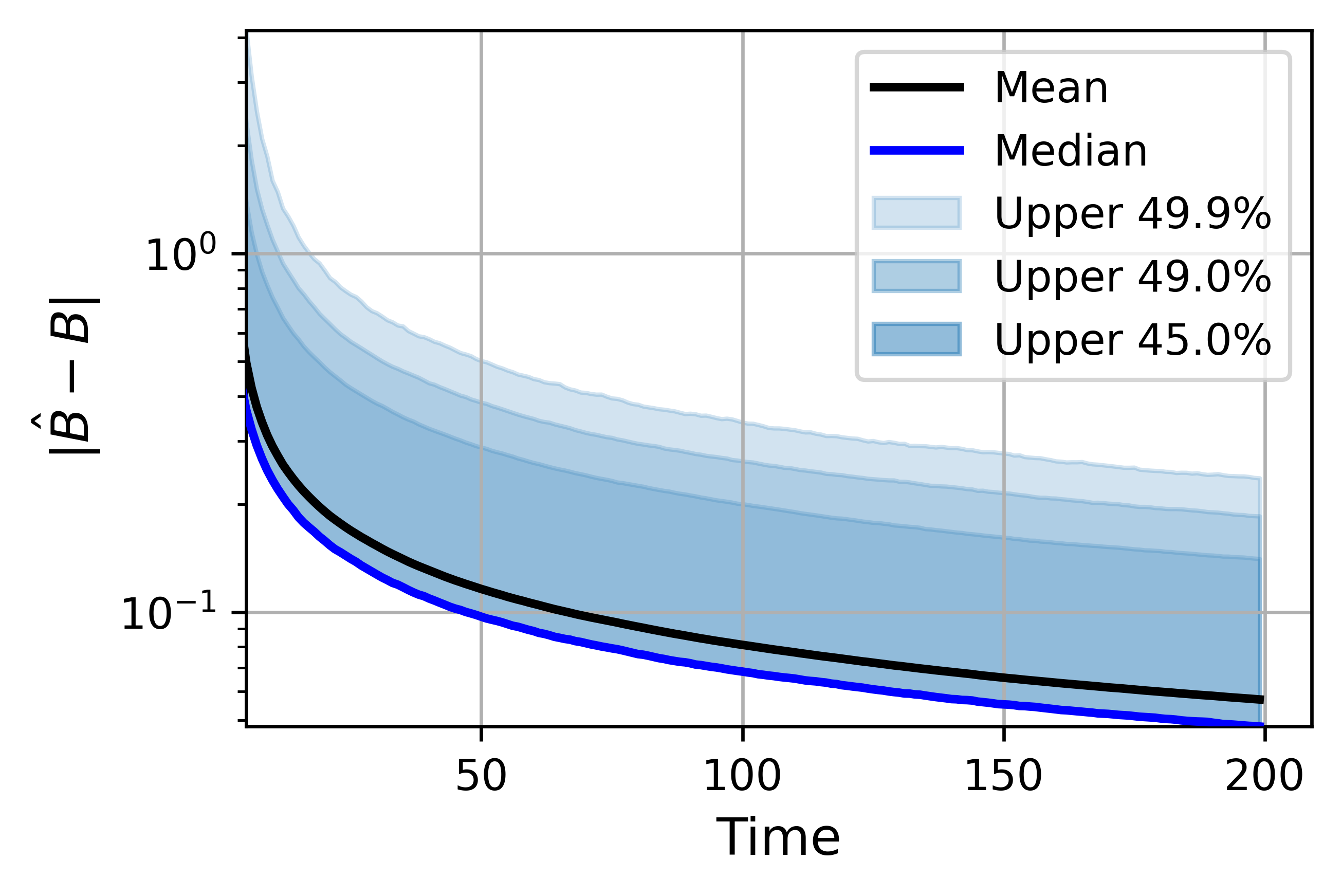}}
  }
\vspace{-0.5cm}
\end{figure}

\begin{figure}[htbp]
\floatconts
  {fig:plot_alphabeta}
  {\vspace{-0.5cm}\caption{(a) State-dependent and (b) control-dependent multiplicative noise variances vs time for the example system using multiplicative noise control.}}
  {%
    \subfigure[$\alpha$]{\label{fig:plot_alpha}%
      \includegraphics[width=0.42\linewidth]{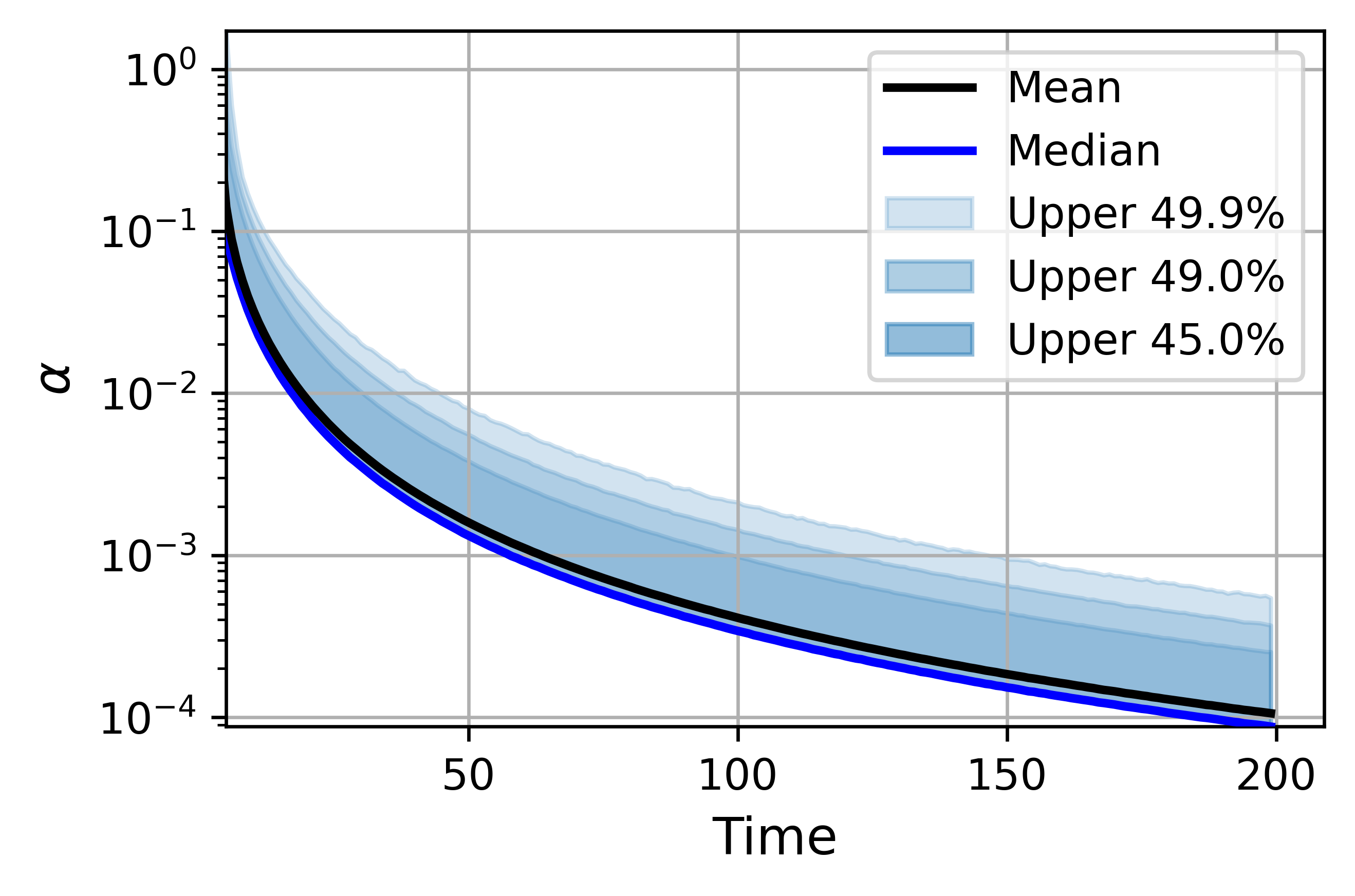}}%
    \qquad
    \subfigure[$\beta$]{\label{fig:plot_beta}%
      \includegraphics[width=0.42\linewidth]{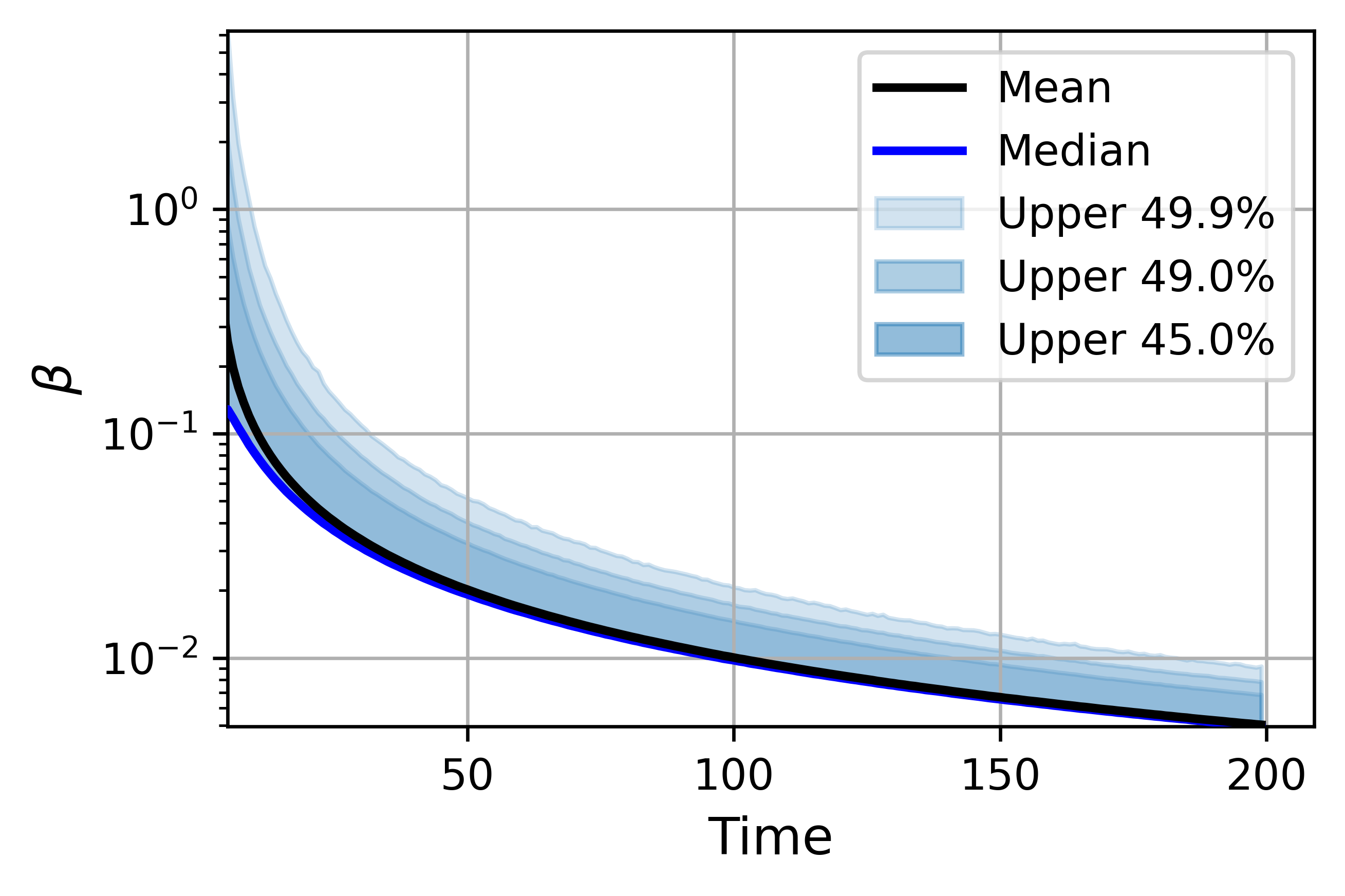}}
  }
\vspace{-0.5cm}
\end{figure}

\begin{figure}[htbp]
    \centering
    \includegraphics[width=0.47\linewidth]{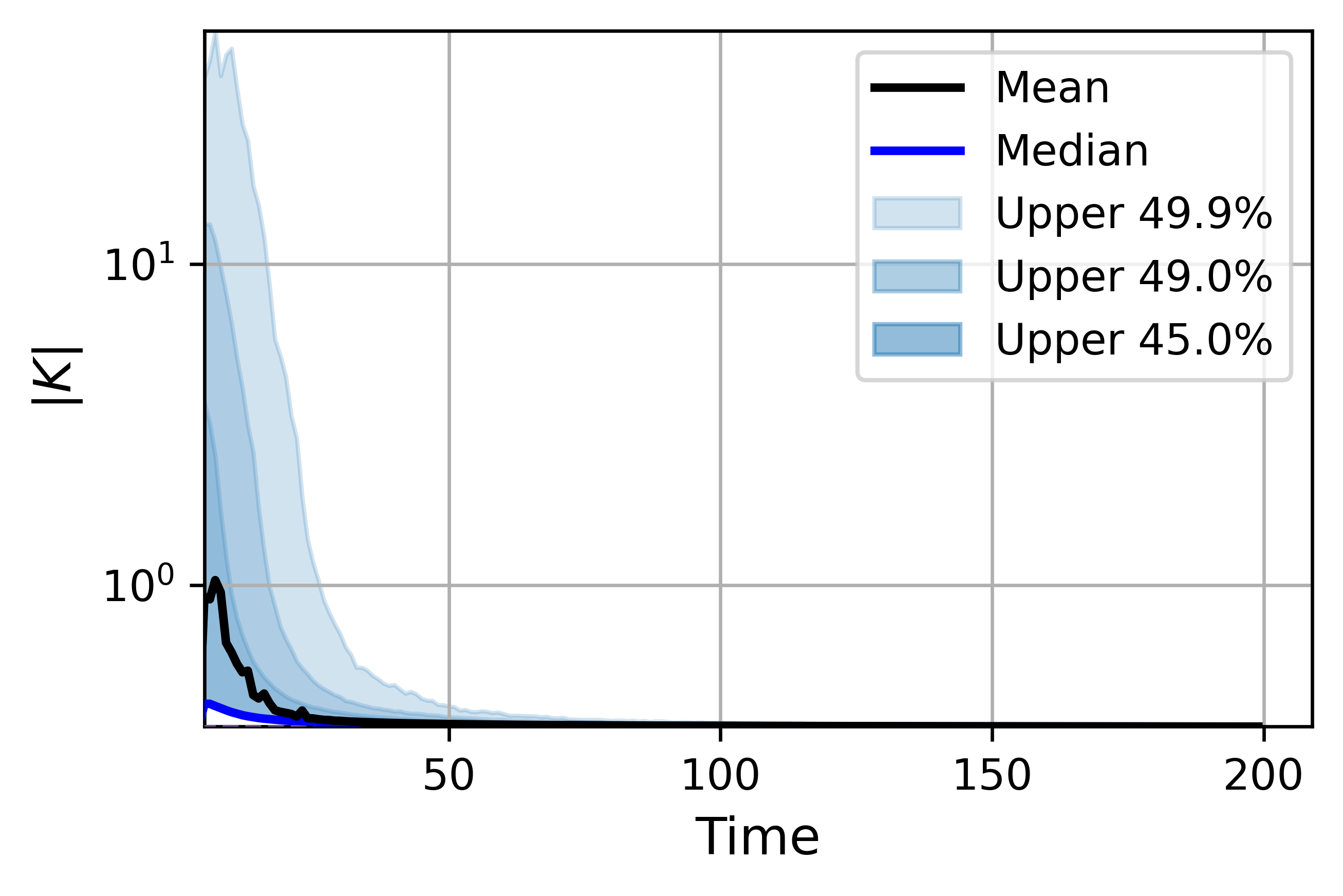}
    \caption{Scaling of multiplicative noise scale parameter $\gamma$ vs time for the example system using multiplicative noise control.}
    \label{fig:plot_gamma_scale}
\end{figure}

\newpage 

\bibliography{refs}

\begin{thebibliography}{17}
\providecommand{\natexlab}[1]{#1}
\providecommand{\url}[1]{\texttt{#1}}
\expandafter\ifx\csname urlstyle\endcsname\relax
  \providecommand{\doi}[1]{doi: #1}\else
  \providecommand{\doi}{doi: \begingroup \urlstyle{rm}\Url}\fi

\bibitem[Abbasi-Yadkori and Szepesv{\'a}ri(2011)]{abbasi2011regret}
Yasin Abbasi-Yadkori and Csaba Szepesv{\'a}ri.
\newblock Regret bounds for the adaptive control of linear quadratic systems.
\newblock In \emph{Proceedings of the 24th Annual Conference on Learning
  Theory}, pages 1--26, 2011.

\bibitem[{\AA}str{\"o}m and Wittenmark(2013)]{aastrom2013adaptive}
Karl~J {\AA}str{\"o}m and Bj{\"o}rn Wittenmark.
\newblock \emph{Adaptive control}.
\newblock Courier Corporation, 2013.

\bibitem[Bellman(1961)]{bellman1961adaptive}
Richard~E Bellman.
\newblock \emph{Adaptive control processes: a guided tour}.
\newblock Princeton university press, 1961.

\bibitem[Bernstein and Greeley(1986)]{bernstein1986robust}
D~Bernstein and S~Greeley.
\newblock Robust controller synthesis using the maximum entropy design
  equations.
\newblock \emph{IEEE Transactions on Automatic Control}, 31\penalty0
  (4):\penalty0 362--364, 1986.

\bibitem[Burden et~al.(1978)Burden, Faires, and Reynolds]{burden1978numerical}
Richard~L Burden, J~Douglas Faires, and Albert~C Reynolds.
\newblock \emph{Numerical analysis}.
\newblock PWS publishing company, 1978.

\bibitem[Chen and Guo(2012)]{chen2012identification}
Han-Fu Chen and Lei Guo.
\newblock \emph{Identification and stochastic adaptive control}.
\newblock Springer Science \& Business Media, 2012.

\bibitem[Dean et~al.(2018)Dean, Mania, Matni, Recht, and Tu]{dean2018regret}
Sarah Dean, Horia Mania, Nikolai Matni, Benjamin Recht, and Stephen Tu.
\newblock Regret bounds for robust adaptive control of the linear quadratic
  regulator.
\newblock In \emph{Advances in Neural Information Processing Systems}, pages
  4188--4197, 2018.

\bibitem[Dean et~al.(2019)Dean, Mania, Matni, Recht, and Tu]{Dean2019}
Sarah Dean, Horia Mania, Nikolai Matni, Benjamin Recht, and Stephen Tu.
\newblock On the sample complexity of the linear quadratic regulator.
\newblock \emph{Foundations of Computational Mathematics}, Aug 2019.

\bibitem[El~Ghaoui(1995)]{ElGhaoui1995}
Laurent El~Ghaoui.
\newblock State-feedback control of systems with multiplicative noise via
  linear matrix inequalities.
\newblock \emph{Systems \& Control Letters}, 24\penalty0 (3):\penalty0
  223--228, 1995.

\bibitem[Gravell et~al.(2019)Gravell, Esfahani, and
  Summers]{gravell2019learning}
Benjamin Gravell, Peyman~Mohajerin Esfahani, and Tyler Summers.
\newblock Learning robust control for {LQR} systems with multiplicative noise
  via policy gradient.
\newblock \emph{CoRR}, 2019.
\newblock URL \url{http://arxiv.org/abs/1905.13547}.

\bibitem[H{\"a}rdle et~al.(2003)H{\"a}rdle, Horowitz, and
  Kreiss]{hardle2003bootstrap}
Wolfgang H{\"a}rdle, Joel Horowitz, and Jens-Peter Kreiss.
\newblock Bootstrap methods for time series.
\newblock \emph{International Statistical Review}, 71\penalty0 (2):\penalty0
  435--459, 2003.

\bibitem[Kumar and Varaiya(2015)]{kumar2015stochastic}
Panqanamala~Ramana Kumar and Pravin Varaiya.
\newblock \emph{Stochastic systems: Estimation, identification, and adaptive
  control}, volume~75.
\newblock SIAM, 2015.

\bibitem[Ljung(1998)]{ljung1998system}
Lennart Ljung.
\newblock \emph{System Identification: Theory for the User}.
\newblock Pearson Education, 1998.

\bibitem[Mania et~al.(2019)Mania, Tu, and Recht]{mania2019certainty}
Horia Mania, Stephen Tu, and Benjamin Recht.
\newblock Certainty equivalence is efficient for linear quadratic control.
\newblock In \emph{Advances in Neural Information Processing Systems}, pages
  10154--10164, 2019.

\bibitem[Milanese and Vicino(1991)]{milanese1991optimal}
Mario Milanese and Antonio Vicino.
\newblock Optimal estimation theory for dynamic systems with set membership
  uncertainty: an overview.
\newblock \emph{Automatica}, 27\penalty0 (6):\penalty0 997--1009, 1991.

\bibitem[Simon(2006)]{simon2006optimal}
Dan Simon.
\newblock \emph{Optimal state estimation}.
\newblock John Wiley \& Sons, 2006.

\bibitem[Wonham(1967)]{Wonham1967}
W~Murray Wonham.
\newblock Optimal stationary control of a linear system with state-dependent
  noise.
\newblock \emph{SIAM Journal on Control}, 5\penalty0 (3):\penalty0 486--500,
  1967.

\end{thebibliography}

\end{document}